\pdfoutput=1

\documentclass[11pt]{article}

\usepackage[final]{acl}

\usepackage{times}
\usepackage{latexsym}

\usepackage[T1]{fontenc}

\usepackage[utf8]{inputenc}

\usepackage{microtype}

\usepackage{inconsolata}

\usepackage{graphicx}
\usepackage{booktabs}
\definecolor{codegreen}{rgb}{0.0, 0.411, 0.243}
\definecolor{codered}{rgb}{0.89, 0.26, 0.20}
\usepackage{hyperref}
\definecolor{dartgreen}{HTML}{00693e}
\definecolor{refcolor}{HTML}{9F363A}
\hypersetup{
    colorlinks=true,
    linkcolor=refcolor,
    citecolor=refcolor,
    filecolor=magenta,      
    urlcolor=refcolor,
    }

\title{\textit{Is It Navajo?}\\Accurate Language Detection in Endangered Athabaskan Languages}

\author{
Ivory Yang \quad 
Weicheng Ma \quad
Chunhui Zhang \quad
Soroush Vosoughi \\
{Department of Computer Science, Dartmouth College} \\
{\small \texttt{\{Ivory.Yang.GR, Weicheng.Ma, Chunhui.Zhang.GR, Soroush.Vosoughi\}@dartmouth.edu}}\\
}

\begin{document}
\maketitle
\linespread{0.90}
\begin{abstract}
Endangered languages, such as Navajo—the most widely spoken Native American language—are significantly underrepresented in contemporary language technologies, exacerbating the challenges of their preservation and revitalization. This study evaluates Google's Language Identification (LangID) tool, which does not currently support any Native American languages. To address this, we introduce a random forest classifier trained on Navajo and twenty erroneously suggested languages by LangID. Despite its simplicity, the classifier achieves near-perfect accuracy (97-100\%). Additionally, the model demonstrates robustness across other Athabaskan languages—a family of Native American languages spoken primarily in Alaska, the Pacific Northwest, and parts of the Southwestern United States—suggesting its potential for broader application. Our findings underscore the pressing need for NLP systems that prioritize linguistic diversity and adaptability over centralized, one-size-fits-all solutions, especially in supporting underrepresented languages in a multicultural world. This work directly contributes to ongoing efforts to address cultural biases in language models and advocates for the development of culturally localized NLP tools that serve diverse linguistic communities.
\end{abstract}

\section{Introduction}
The urgency of preserving endangered languages is not merely a linguistic issue but one deeply connected to the preservation of cultural, historical, and ecological knowledge \cite{tulloch2006preserving, zariquiey-etal-2022-cld2, zhang-etal-2022-nlp, cusenza2024nlp, yang2025nushurescue}. These languages reflect the intellectual heritage of diverse communities, playing a critical role in maintaining global cultural diversity. Yet, despite this significance, the development of language technologies has been disproportionately focused on languages with large speaker bases and economic clout, leaving languages with smaller populations—such as Native American languages—largely unsupported.

This study centers on Navajo, the most widely spoken Native American language \cite{dietrich2022language}, which remains unsupported by commercial language technologies like Google's Language Identification (LangID) tool \cite{caswell2020language}. The lack of comprehensive linguistic datasets and dedicated tools impedes both language preservation and learning efforts \cite{shamsfard2019challenges}. This technological gap is even more pronounced for other Native American languages, many of which are on the verge of extinction due to minimal technological integration and educational resources \cite{meredith2013racing, flavelle2023strengthening}.

\begin{figure}[t]
  \centering
  \includegraphics[width=0.95\linewidth]{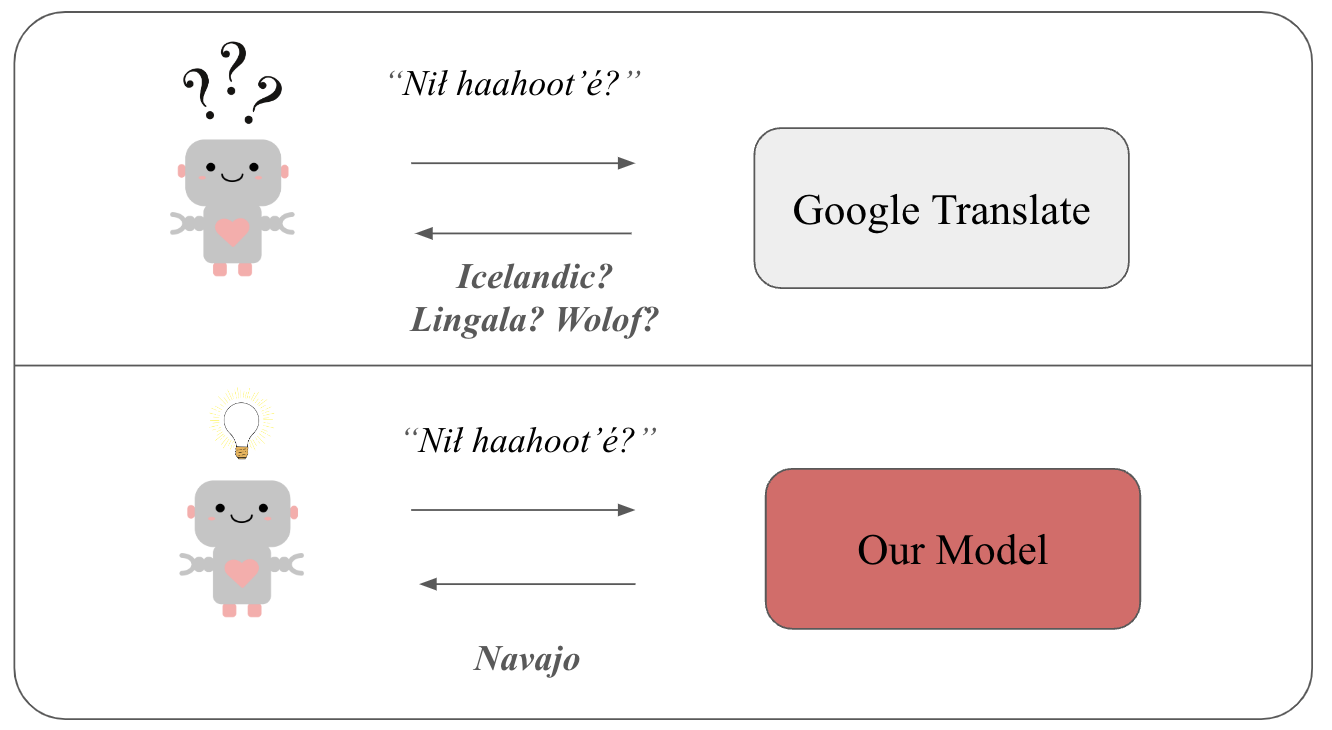}
  \caption{Google's LangID does not currently support any Native American languages, including Navajo and other Athabaskan languages. Our model addresses these challenges effectively.}
  \label{fig:intropic}
  \vspace{-9pt}
\end{figure}

Google LangID's performance on the \textit{Navajo 10k} dataset \cite{goldhahn2012building} revealed complete misidentification of Navajo sentences as unrelated languages, an expected outcome given that LangID does not currently support any Native American language. In response, we developed a language identification model tailored to accurately distinguish Navajo from languages erroneously suggested by LangID, achieving near-perfect accuracy. This success illustrates that low-resource languages, often overlooked by major technological platforms, can be effectively supported with targeted approaches and resources. Beyond Navajo, we extended our model to other languages in the  \textbf{Athabaskan family}—including \textit{Western Apache, Mescalero Apache, Jicarilla Apache, and Lipan Apache} \cite{georgecultural}. Our model’s robustness across these related languages underscores its potential applicability across broader linguistic groups (see Figure \ref{fig:intropic}). This suggests a viable path for NLP technologies to not only support individual endangered languages but to facilitate revitalization efforts across entire language families.

\textbf{We highlight how centralization in language technology disproportionately benefits global languages, often sidelining underrepresented languages and thus contributing to the erosion of linguistic diversity \cite{schneider2022multilingualism}}. Our findings underscore the feasibility of creating decentralized, robust language identification tools, which, by focusing on the unique needs of specific languages, can play a significant role in preserving endangered languages. Furthermore, it offers promising pathways for leveraging NLP tools across culturally and linguistically related groups, enriching both academic research and community-driven language revitalization by fostering tools that are responsive to the specific needs of these communities. This aligns with the broader goal of developing NLP technologies that not only accommodate but also actively support the linguistic and cultural diversity of our vibrant multicultural world.

\section{Background}
While there exist studies on endangered languages~\cite{zariquiey-etal-2022-cld2, zhang-etal-2022-nlp, cusenza2024nlp}, their integration into business technologies remains insufficient. For example, although Google's LangID supports over a hundred languages, it fails to include any Native American language, and so provides completely inaccurate suggestions when encountering Navajo. Similarly, advanced NLP models, such as ChatGPT, struggle with Navajo due to a lack of training data, which is predominantly derived from more widely spoken languages~\cite{hangya2022improving}.

The scarcity of digital resources for Navajo further compounds these challenges, as it lacks sufficient digital presence needed for effective NLP tool development \cite{magueresse2020low}. This scarcity not only limits the use of standard NLP methodologies but also hampers preservation and revitalization efforts.
These issues reflect broader market-driven priorities in language technology, which overlook less commercially viable languages \cite{costa2022no}, creating significant barriers to preserving cultural heritage and emphasizing the need for more inclusive technological support for endangered languages \cite{todacheeny2014navajo}.

\section{Identification for NatAm Languages}
\begin{figure}[t]
  \centering
  \includegraphics[width=\linewidth]{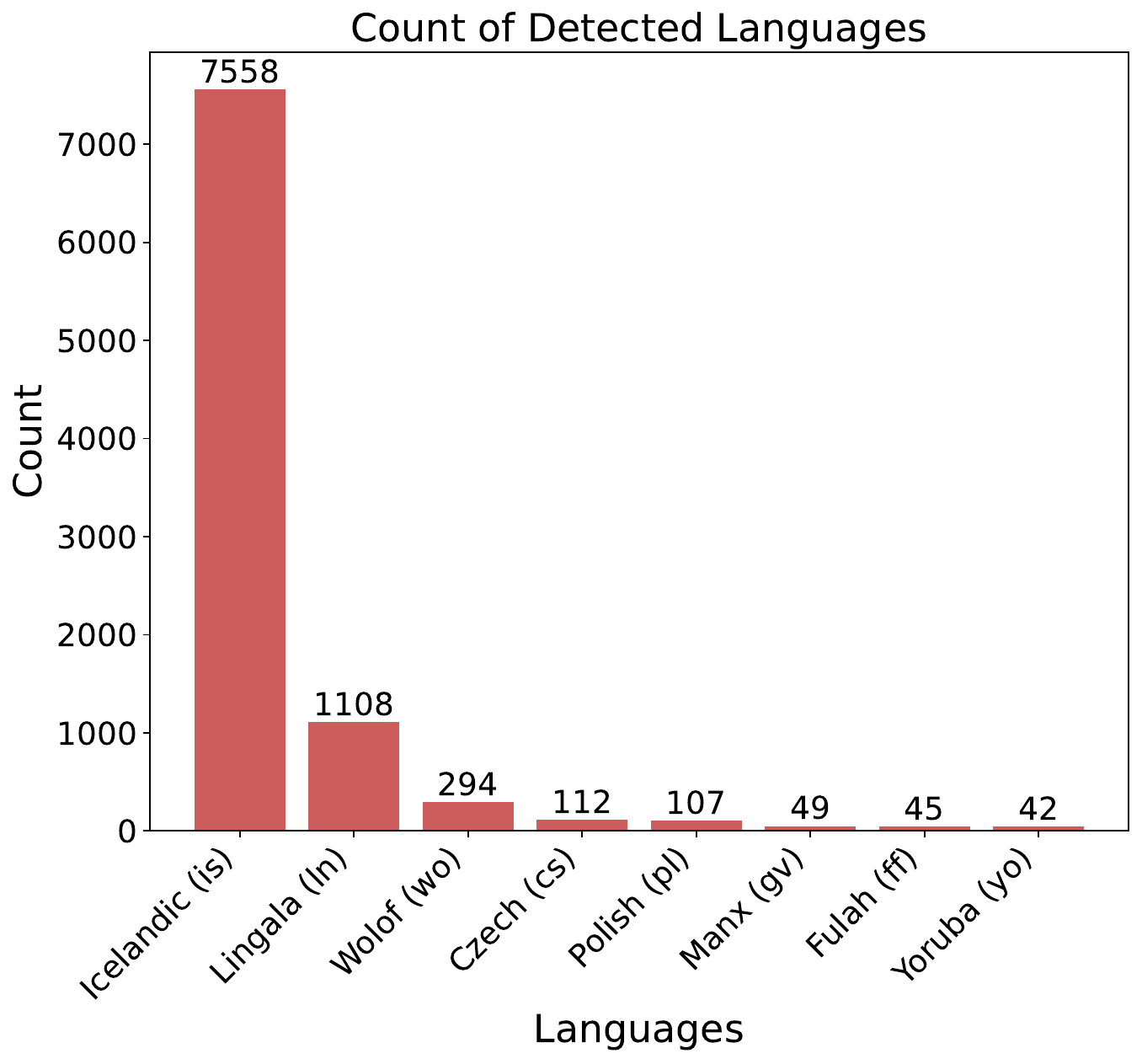}
  \vspace{-0.3in}
  \caption{Visual representation of some erroneously suggested languages by Google LangID, along with their frequency counts.}
  \label{fig:lfreq}
    \vspace{-10pt}
\end{figure}
The benefits of Native American language identification are twofold: to bolster the development of linguistic tools tailored to these languages, and to aid in their preservation and revitalization \cite{mohanty2023role}. Effective identification is foundational for creating technologies that understand and process these languages, addressing the significant digital divide in language technology support \cite{mohanty2024applying}. Our evaluation aims at a detailed assessment of the models' capability to accurately recognize and differentiate between Native American languages and others.
By understanding the strengths and limitations of our models, we can refine our techniques to better serve the needs of Native American language communities.

\subsection{Benchmark Construction}
To construct our dataset for evaluating language identification models, we used two distinct approaches to account for diversity and specificity. The first dataset was formed based on twenty languages\footnote{The languages are Icelandic, Lingala, Wolof, Czech, Polish, Manx, Fulah, Yoruba, Portuguese, Somali, Slovak, Tsonga, Spanish, Oromo, Indonesian, Igbo, Northern Sami, Irish, Arabic and English.} that Google's LangID misidentifies as when presented with Navajo sentences, with their distribution shown in Figure \ref{fig:lfreq}. Each entry consists of 1k or 10k sentences extracted from the Leipzig corpora \cite{goldhahn2012building}, ensuring standardization against the Navajo 10k dataset in terms of linguistic features and contexts.

The second dataset focuses on languages from the Athabaskan family, which are linguistically related to Navajo, including Western Apache, Mescalero Apache, Jicarilla Apache, and Lipan Apache~\cite{saxon202339}. A sample of aligned words across these languages are included in Appendix \ref{sec:appendix}. To compile this dataset, we curated a limited sample of texts contributed by native speakers and language preservation organizations. Specifically, we collected texts in Western Apache~\cite{glosbe2024}, Mescalero Apache~\cite{mescalero2024}, Jicarilla Apache~\cite{jicarilla2024}, and Lipan Apache~\cite{lipan2024}. These curated texts not only enrich the model's exposure to authentic linguistic scenarios but also enable the evaluation of its ability to discern subtle linguistic nuances among closely related languages, thereby enhancing the precision and applicability of our language identification models.

\begin{figure}[t]
  \centering
  \includegraphics[width=\linewidth]{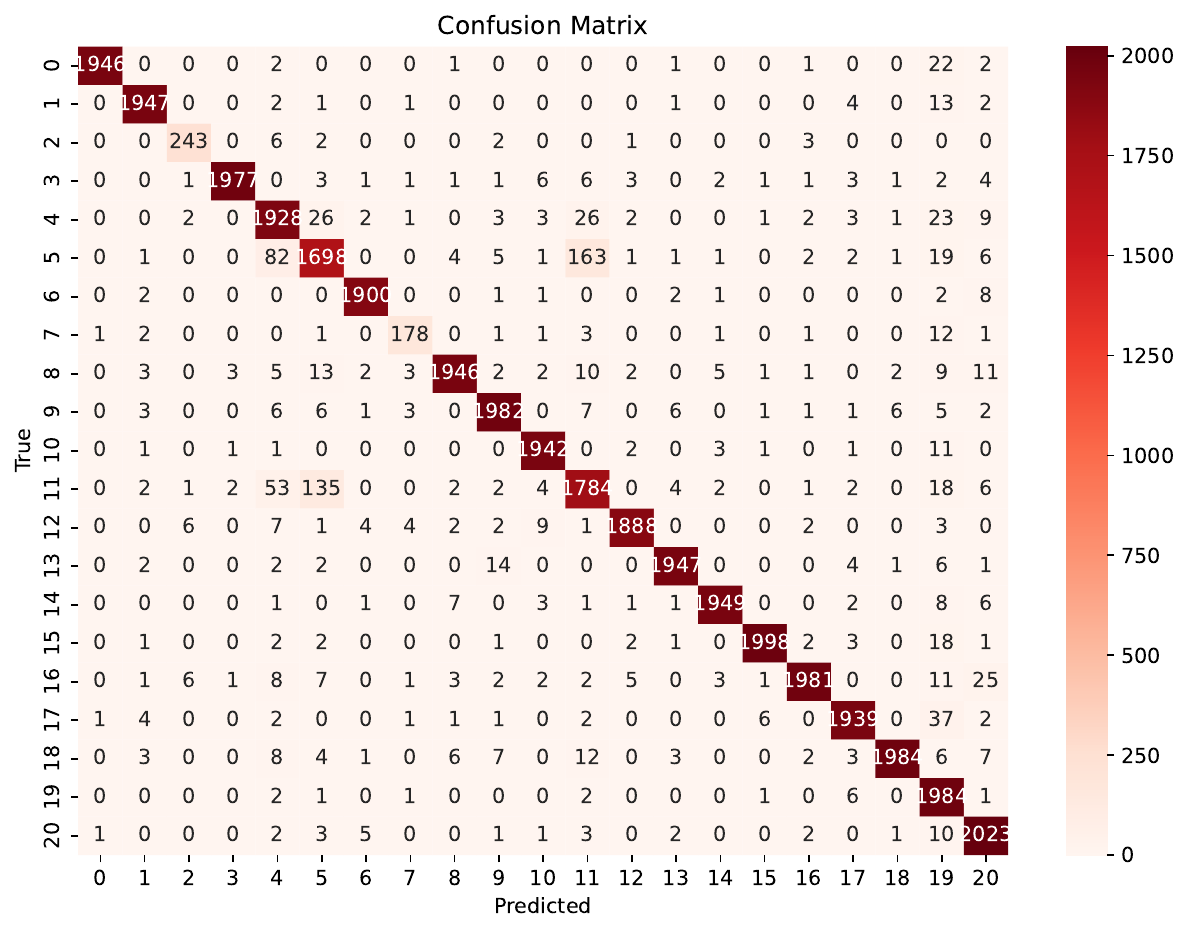}
  \caption{Classification results for Navajo and 20 other languages, presented as a confusion matrix. The intensity of the color indicates the frequency of classifications. Labels on x and y axes correspond to the languages listed in Table \ref{tbl:classification_metrics}.}
  \label{fig:cm}
\end{figure}

\subsection{Model Specifics}
For the language identification task, we used a Random Forest classifier \cite{parmar2019review} with 100 trees for its robustness, interpretability, and effectiveness in handling complex classification tasks. The model builds multiple decision trees and predicts by aggregating their results, which reduces overfitting and enhances generalization. This model was chosen for its high interpretability, crucial when working with lesser-known languages, as it helps identify key features that distinguish languages, supporting linguistic analysis and improving endangered language identification. Additionally, its simplicity and widespread use make it ideal for replication and comparison with similar research.

\begin{table}
  \centering
    \resizebox{0.485\textwidth}{!}{ 
  \begin{tabular}{cccccc}
    \toprule
    \textbf{Language} & \textbf{Class} & \textbf{Precision} & \textbf{Recall} & \textbf{F1-score} & \textbf{Support} \\
    \midrule
    Navajo & 0 & 1.00 & 0.99 & 0.99 & 1975 \\
    Icelandic & 1 & 0.99 & 0.99 & 0.99 & 1971 \\
    Lingala & 2 & 0.94 & 0.95 & 0.94 & 257 \\
    Wolof & 3 & 1.00 & 0.98 & 0.99 & 2014 \\
    Polish & 4 & 0.91 & 0.95 & 0.93 & 2032 \\
    Czech & 5 & 0.89 & 0.85 & 0.87 & 1987 \\
    Manx & 6 & 0.99 & 0.99 & 0.99 & 1917 \\
    Fulah & 7 & 0.92 & 0.88 & 0.90 & 202 \\
    Yoruba & 8 & 0.99 & 0.96 & 0.97 & 2020 \\
    Portuguese & 9 & 0.98 & 0.98 & 0.98 & 2030 \\
    Somali & 10 & 0.98 & 0.99 & 0.99 & 1963 \\
    Slovak & 11 & 0.88 & 0.88 & 0.88 & 2018 \\
    Tsonga & 12 & 0.99 & 0.98 & 0.98 & 1929 \\
    Spanish & 13 & 0.99 & 0.98 & 0.99 & 1979 \\
    Oromo & 14 & 0.99 & 0.98 & 0.99 & 1980 \\
    Indonesian & 15 & 0.99 & 0.98 & 0.99 & 2031 \\
    Igbo & 16 & 0.99 & 0.96 & 0.98 & 2059 \\
    Northern Sami & 17 & 0.98 & 0.97 & 0.98 & 1996 \\
    Irish & 18 & 0.99 & 0.97 & 0.98 & 2046 \\
    Arabic & 19 & 0.89 & 0.99 & 0.94 & 1998 \\
    English & 20 & 0.96 & 0.98 & 0.97 & 2054 \\
    \bottomrule
  \end{tabular}}
  \caption{\label{tbl:classification_metrics}
    Detailed classification results for each language: precision, recall, f1-score, and support counts.
  }
\end{table}

\subsection{Experimental Results}
Our experimental evaluation of the Random Forest classifier is conducted using a dataset that included Navajo, labeled as '0', alongside a selection of languages that Google's LangID erroneously identifies as similar to Navajo. The dataset comprised 153,832 training samples and 38,458 test samples, each vectorized into 5,000 features to capture a wide range of linguistic attributes. The classifier achieved an overall accuracy of 97\%, demonstrating its strong capability to distinguish Navajo from the misidentified languages.

The performance metrics show high precision, recall, and f1-scores across all tested languages, with Navajo achieving a precision of 1.00, a recall of 0.99, and an f1-score of 0.99, as shown in Table \ref{tbl:classification_metrics}. These results highlight the classifier’s effectiveness in accurately differentiating Navajo from other unrelated languages, which are erroneously suggested by Google's LangID. The confusion matrix in Figure \ref{fig:cm} highlights the classifier's performance, particularly for Navajo. The model achieved 1,946 true positives for Navajo, with only a few false negatives. These results demonstrate the model's robustness in identifying Navajo and underscore its potential for supporting language identification in underrepresented Native American languages. Nevertheless, addressing the few misclassifications through enhanced training could further improve accuracy and generalization.

The stable performance of our classifier serves as a counterresponse to Google's LangID, and its lack of support for Native American languages, including Navajo. This omission directly results in the erroneous suggestions of linguistically unrelated languages, highlighting a critical need to include these languages in global technology platforms to better respect and reflect linguistic diversity.

\subsection{Model Generalizability}

\begin{figure}[t]
  \centering
  \includegraphics[width=\linewidth]{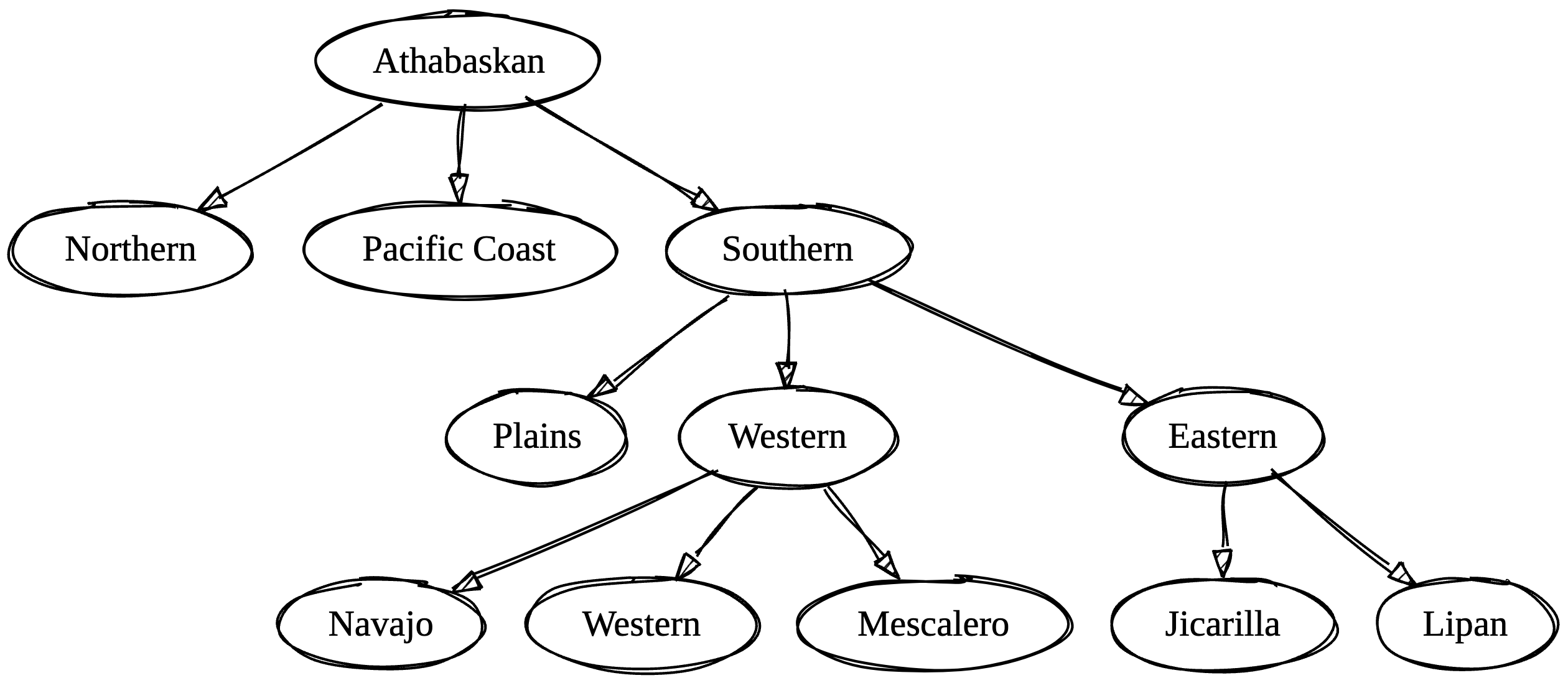}
  \caption{Family Tree for Athabaskan Languages}
  \label{fig:ftree}
\end{figure}

Following the successful differentiation of Navajo from languages erroneously suggested by Google's LangID, we further tested the classifier’s capability with our second dataset of curated Apache languages. Upon running this subset through the classifier, initially trained to distinguish Navajo from other languages, we observed that the classifier often identified these Apache languages as Navajo. This result is particularly significant given the linguistic similarities within the Athabaskan language family, to which both Navajo and the Apache languages belong. The classifier’s performance here underscores its ability not only to identify Navajo with high accuracy but also to generalize across related languages within the same family. This generalizability is indicative of the model's potential utility in broader linguistic applications, especially in creating tools that support multiple but related Native American languages.

These findings also raise interesting questions about the classifier’s sensitivity to the nuances between closely related languages and its potential role in developing more sophisticated NLP tools that can accurately differentiate between languages with subtle linguistic differences. The detection for Navajo performed best for Western Apache and Mescalero Apache, as shown in Table \ref{tbl:apache_classification}. Both these languages fall under the Western Apachean subgroup along with Navajo, as shown in Figure \ref{fig:ftree}. On the other hand, Jicarilla Apache and in particular, Lipan Apache, performed less well in Navajo detection, which could be because they fall under the Eastern Apachean subgroup. This observation could be pivotal for linguistic preservation, allowing for the development of specialized educational and communicational tools tailored to each language's unique characteristics.

\begin{table}
  \centering
    \resizebox{0.485\textwidth}{!}{ 
  \begin{tabular}{@{}lcc@{}}
    \toprule
    \textbf{Language}           & \textbf{Classified as Navajo} & \textbf{Total Sentences} \\
    \midrule
    Western Apache     & 96.00\%         & 25 \\
    Mescalero Apache   & 100.00\%        & 32 \\
    Jicarilla Apache   & 92.31\%         & 13 \\
    Lipan Apache       & 62.16\%         & 37 \\
    \bottomrule
  \end{tabular}}
  \vspace{-0.1in}
  \caption{\label{tbl:apache_classification}
    Classification Results for Apache Languages: Percentage of sentences classified as Navajo and total number of sentences examined for each type of Apache language (out of 107 sentences).
  }
  \vspace{-8pt}
\end{table}

\section{Conclusion and Future Work}
This study demonstrates the effectiveness of our Random Forest classifier in accurately distinguishing Navajo from languages erroneously suggested by Google's LangID, as well as effectively recognizing related Athabaskan languages. These results emphasize the potential for broader applications in language identification, particularly for underrepresented languages. Our findings highlight a significant gap in support for Native American languages in current digital platforms, and urge the need for refined, inclusive language models.

Future work can focus on expanding the classifier's training to include additional Native American languages, improving its adaptability, and extending its utility to different language groups. The development of tools capable of distinguishing closely related languages is crucial for supporting educational and communication needs within Native communities\footnote{This study represents a preliminary exploration, and we acknowledge the importance of direct collaboration with Native American communities. Moving forward, we plan to engage with community members and linguistic experts to ensure our work aligns with their perspectives, priorities, and cultural considerations.}. We also advocate the decentralization of NLP research efforts, emphasizing the need for targeted investment in endangered languages. Such initiatives are essential to ensure that advances in language technology promote linguistic equity, thereby preserving cultural diversity and heritage in the digital age.
\section*{Limitations}
While the study successfully demonstrates the Random Forest classifier's efficacy in distinguishing Navajo from languages commonly misidentified by Google Translate and identifying related Athabaskan languages, it does have limitations that impact its broader applicability. Firstly, the language variety included in the study is limited; the classifier was tested primarily against a small set of languages suggested by Google's LangID and a few Athabaskan languages. This narrow scope might not capture the classifier’s effectiveness across a broader range of Native American languages, potentially limiting its utility for other endangered language families. Secondly, the experimental design assumes a binary distinction between Navajo and other languages without considering intra-group variations and dialectical differences within the Athabaskan language family, which could affect accuracy in real-world applications. Lastly, reliance on vectorized features of 5,000 dimensions may overlook some finer linguistic nuances, which are crucial for distinguishing between closely related languages. Addressing these limitations in future work will be essential for developing more robust and applicable language identification systems.

\section*{Ethics}
Ethical considerations are paramount in the development of language technology, especially for Native American languages, which are deeply intertwined with cultural identity and heritage. This study emphasizes the importance of respectful engagement with these communities, recognizing the cultural, spiritual, and historical significance of their languages. Technology development involving Native American languages should proceed with close collaboration with native speakers and community leaders to ensure that these tools support and reinforce language preservation rather than contributing to cultural homogenization or appropriation. Additionally, data privacy and consent are critical, as much of the linguistic data involves sensitive cultural content. Ensuring that communities retain control over how their linguistic resources are used is essential for maintaining trust and upholding ethical standards in research. Moreover, to ensure transparency and foster research, we have made our code and datasets used publicly available at \url{https://github.com/ivoryayang/Isitnavajo}.

\section*{Acknowledgment}
This work was partially funded by a Google Research Award, and in part by a Dartmouth Alumni Research Award. We extend our gratitude to the Dartmouth graduate alumni for their generous support and commitment to fostering academic inquiry.

\bibliography{references}

\cleardoublepage

\appendix
\begin{figure*}[!h]
  \centering
  \includegraphics[width=0.78\linewidth]{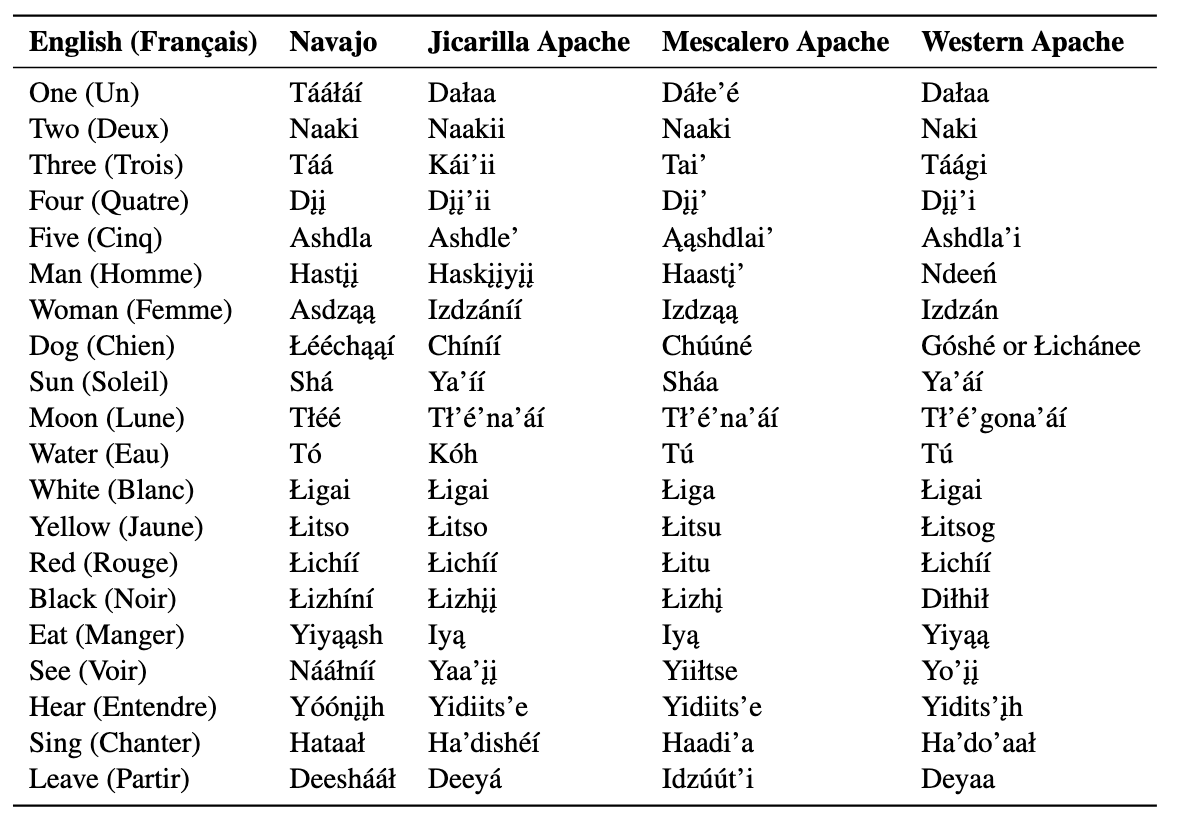}
  \caption{Aligned Words from \citet{apache}.}
  \label{fig:appendix}
\end{figure*}
\section{Aligned Words Across Native American Languages} \label{sec:appendix}
Figure \ref{fig:appendix} lists 20 aligned words in four Native American languages, together with their English and French translations.

\end{document}